\documentclass[lettersize,journal]{IEEEtran}
\usepackage{cite}
\usepackage{amsmath,amssymb,amsfonts}
\usepackage{algorithmic}
\usepackage{graphicx}
\usepackage[caption=false, font=footnotesize]{subfig}
\usepackage{lipsum}
\usepackage{textcomp}
\usepackage{verbatim}
\usepackage{booktabs,caption}
\usepackage{multirow}
\usepackage[T1]{fontenc}
\usepackage{algorithm}
\usepackage{color}
\usepackage{siunitx}
\usepackage{threeparttable}
\usepackage{marvosym}
\usepackage{makecell} 
\usepackage{xspace}
\newcommand{\ie}{\textit{i}.\textit{e}.}
\newcommand{\eg}{\textit{e}.\textit{g}.}
\newcommand{\etal}{\textit{et al}.}

\usepackage[colorlinks,hyperindex,breaklinks]{hyperref}
\begin{document}
\title{BoNuS: Boundary Mining for Nuclei Segmentation with Partial Point Labels}
\author{Yi~Lin, 
        Zeyu~Wang, 
        Dong~Zhang, \IEEEmembership{Member, IEEE,} \\
        Kwang-Ting~Cheng, \IEEEmembership{Fellow, IEEE},
        and Hao~Chen, \IEEEmembership{Senior Member, IEEE}
\thanks{Y. Lin, Z. Wang, D. Zhang, and K.-T. Cheng are with the Department of Computer Science and Engineering, The Hong Kong University of Science and Technology, Hong Kong, China. E-mail: \{yi.lin, zwangek\}@connect.ust.hk; \{dongz, timcheng\}@ust.hk;}
\thanks{H. Chen is with the Department of Computer Science and Engineering and Department of Chemical and Biological Engineering, Hong Kong University of Science and Technology, Hong Kong, China.} 
\thanks{H. Chen is also affiliated with HKUST Shenzhen-Hong Kong Collaborative Innovation Research Institute, Futian, Shenzhen, China. E-mail: E-mail: jhc@cse.ust.hk}
\thanks{Y. Lin and Z. Wang contribute equally; Corresponding author: H. Chen.}
}
\maketitle
\begin{abstract}
Nuclei segmentation is a fundamental prerequisite in the digital pathology workflow. 
The development of automated methods for nuclei segmentation enables quantitative analysis of the wide existence and large variances in nuclei morphometry in histopathology images.
However, manual annotation of tens of thousands of nuclei is tedious and time-consuming, which requires significant amount of human effort and domain-specific expertise.
To alleviate this problem, in this paper, we propose a weakly-supervised nuclei segmentation method that only requires partial point labels of nuclei.
Specifically, we propose a novel boundary mining framework for nuclei segmentation, named BoNuS, which simultaneously learns nuclei interior and boundary information from the point labels. 
To achieve this goal, we propose a novel boundary mining loss, which guides the model to learn the boundary information by exploring the pairwise pixel affinity in a multiple-instance learning manner.
Then, we consider a more challenging problem, i.e., partial point label, where we propose a nuclei detection module with curriculum learning to detect the missing nuclei with prior morphological knowledge.
The proposed method is validated on three public datasets, MoNuSeg, CPM, and CoNIC datasets.
Experimental results demonstrate the superior performance of our method to the state-of-the-art weakly-supervised nuclei segmentation methods.
Code: \url{https://github.com/hust-linyi/bonus}.
\end{abstract}


\section{Introduction}
\label{sec:introduction}
Histopathology slides contain a wealth of phenotypic information and play a vital role in cancer diagnosis, prognosis, and treatment~\cite{elmore2015diagnostic}.
Through the staining of agents such as hematoxylin and eosin, the tissue structure and nuclei morphology can be visualized and analyzed by pathologists, which is important or even sufficient for the diagnosis of many cancers.
With the development of computational pathology, automatic analysis of H\&E stained histopathology slides has become a hot topic in the field of medical image analysis, which can help pathologists reduce the workload and improve the efficiency of diagnosis.
Nuclei segmentation is a critical step in the analysis of histopathology slides, and it serves as the basis for the subsequent analysis and management of the tissue specimens and whole slide images~\cite{graham2019hover}. 
For example, the nuclei segmentation results can be used to estimate the tumor cellularity, which is an important indicator for the prognosis of cancer patients~\cite{chang2012invariant}.
Therefore, automatic nuclei segmentation methods have been widely studied and play an important role in the analysis of histopathology slides.
\begin{figure}
\centering
\includegraphics[width=0.48\textwidth]{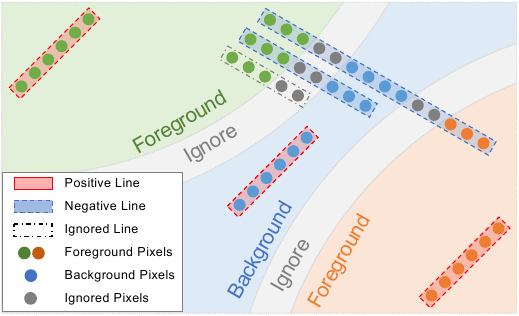}
\caption{Illustration for boundary loss, where the boundary prediction can be supervised by pixel-affinity via multiple instance learning.}
\label{fig:mil}
\end{figure}


Recent years have witnessed the tremendous success of deep learning in medical image segmentation~\cite{kumar2017dataset,chen2017dcan,zhou2019cia}.
However, as a data-hungry method, state-of-the-art deep learning methods for nuclei segmentation usually require a large number of annotated nuclei images.
Manual annotation of nuclei images is time-consuming and labor-intensive, making it expensive and impractical for histopathology slide analysis.
Further, with the annotation process continuing, manual annotation accuracy and speed will gradually decrease, due to the attention span and eye fatigue of the pathologists~\cite{jara2010digital}.
Weak annotations, such as bounding boxes or point annotations, are more practical and efficient, which could save the annotation time by 42\% and 88\%~\cite{qu2020weakly}, respectively.
Therefore, intensive research has been conducted on weakly supervised nuclei segmentation methods with point annotation to balance the model performance and annotation efficiency. 
One of the main challenges of this task is how to effectively utilize the point annotation information to improve the segmentation performance.
Existing methods typically inject the domain knowledge of nuclei into the segmentation model.
For example, using the shape and size prior of nuclei to generate supervision signals, such as Gaussian heatmaps~\cite{alom2018microscopic}, Voronoi diagrams~\cite{brown1979voronoi}, and distance maps~\cite{naylor2018segmentation}.
Although these methods can effectively ease the annotation burden, they are still limited by the incomplete and inaccurate generated supervision signals, especially for the nuclei boundary.

In this paper, we propose BoNus, a novel \textbf{bo}undary mining framework for \textbf{nu}clei \textbf{s}egmentation.
In contrast to the existing methods that implicitly refine the segmentation results for the nuclei boundary, such as self-training~\cite{xie2020instance}, co-training~\cite{lin2023nuclei}, and self-supervised learning~\cite{xie2020instance}, BoNuS explicitly mines the boundary information from the point annotation.
Specifically, we first introduce a novel boundary mining loss to guide the model to learn the boundary information from the point annotations.
We address the unavailability of the boundary information by formulating this problem as a multiple instance learning (MIL) problem and integrating the proposed MIL formulation into fully supervised boundary prediction.
In this way, the boundary information, as well as the instance-aware feature representation, can be learned from the pseudo-labels generated by the coarse-stage segmentation model, which is trained by the incomplete and inaccurate supervision signals (\eg, Voronoi and cluster labels).

The idea behind BoNuS is inspired by the recent success of the MIL framework in the field of weakly supervised image segmentation.
MIL is a learning paradigm where the training data are composed of a set of instances and bags, each bag contains a set of instances. The goal of MIL is to learn a classifier that predicts the label of a bag based on the labels of its instances. This is according to the concept that a bag is positive if it contains at least one positive instance, and negative otherwise.
As shown in Fig.~\ref{fig:mil}, a nuclei segmentation map categorizes each pixel into one of the three classes: background, foreground, and ignored area.
In our MIL formulation, the line connecting any two pixels in the segmentation map is considered as a bag, and the pixels along the line are considered as instances. 
A line is positive if the pixels along the line are all background or in the same nucleus, otherwise it is negative.
We also ignore the lines whose endpoints are in the ignored area.
In other words, a line would cross the boundary of a nucleus if its two ends are in different classes or nuclei.
We integrate the MIL formulation into the nuclei segmentation model and propose a novel boundary mining loss to explicitly guide the model to learn the boundary information from the point annotation.

The main contributions of this paper are summarized as follows. 
\emph{First}, we propose BoNuS, a novel boundary mining framework for nuclei segmentation. 
The proposed framework can simultaneously learn the interior and boundary information from the point annotation, which is more accurate and efficient than the existing methods.
\emph{Second}, we propose a novel boundary mining loss to explicitly guide the model to learn the boundary information by exploring the pairwise pixel affinities in a MIL formulation.
\emph{Third}, we tackle a more challenging task, partial point annotation, where only portions of the nuclei are annotated. We propose a nuclei detection module that integrates the domain knowledge of nuclei into the detection module with curriculum learning and uses the detection results to generate the point annotation for the segmentation model.
\emph{Finally}, we conduct extensive experiments on the MoNuSeg~\cite{monuseg_test}, CPM~\cite{cpm}, and CoNIC~\cite{graham2021conic} datasets, and demonstrate the effectiveness of each component of our proposed framework and superior performance compared with state-of-the-art methods. 
\begin{figure*}
\centering
\includegraphics[width=\textwidth]{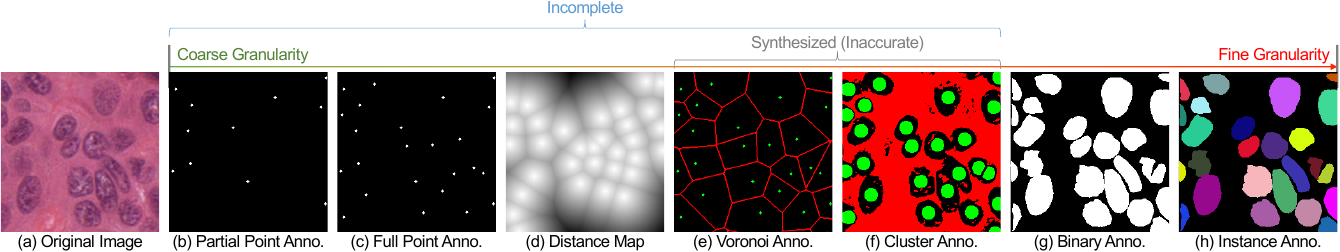}
\caption{This Figure illustrates the annotations at different granularities for nuclei image segmentation. The original image is shown in (a), followed by partial point annotations in (b) and full point annotations in (c). The distance transformation map is shown in (d), while Voronoi annotations and cluster annotations are synthesized from the full point annotations and displayed in (e) and (f), respectively. Pixel-wise binary annotations and instance annotations are shown in (g) and (h), respectively. It should be noted that the synthesized Voronoi and cluster annotations may be inaccurate and incomplete. The foreground, background, and ignore areas are highlighted in \textcolor{green}{green}, \textcolor{red}{red}, and black, respectively.}
\label{fig:labels}
\end{figure*}
\section{Related Work}
\label{sec:related_work}
\subsection{Nuclei Detection and Segmentation}
\label{sec:related_work_nuclei_detection}
Segmentation of nuclei in pathology images holds significant clinical value, as it provides essential morphological and genetic information that can aid in the diagnosis of various diseases, including several types of cancers.
Among the methods used for nuclei segmentation, energy-based techniques, particularly the watershed algorithm\cite{yang2006nuclei}, have been widely employed to automatically detect and segment nuclear instances.
However, the watershed algorithm is sensitive to thresholds and the initial markers, which may result in inaccurate segmentation results. 
To address these issues, active contour approaches have been proposed to obtain better markers\cite{ali2012integrated}, and the energy landscape has been improved by incorporating the morphological information of nuclei~\cite{liao2016automatic}. 
Despite these advancements, predefined hand-crafted features may not be suitable for all nuclei images, and the quality of the initial markers still poses a limitation to the performance of these methods.

With the recent advances in deep learning, numerous methods have been proposed for the automatic detection and segmentation of nuclei in medical images. For instance, Micro-Net~\cite{raza2019micro} extended the U-Net~\cite{ronneberger2015u} by extracting multi-scale features to detect and segment nuclei of varying sizes. NuClick~\cite{koohbanani2020nuclick} proposed a semi-automatic segmentation method with squiggles as the guiding signal. Furthermore, several methods have been proposed for the instance segmentation of nuclei in medical images. Chen~\etal~\cite{chen2017dcan} proposed a deep contour-aware network that separately learns the contours and the interior of nuclei to split the touching nuclei. Based on the concept of using nuclear contours, Kumar~\etal~\cite{kumar2017dataset} built a nuclei instance segmentation dataset using deep learning to automatically annotate the contours of nuclei. Zhou~\etal~\cite{zhou2019cia} designed a contour-aware information aggregation network to integrate the contour and interior information of nuclei to improve the segmentation performance. Additionally, several methods have proposed to use other supervision information to enhance the nuclei instance segmentation performance, such as distance map~\cite{naylor2018segmentation} and displacement field~\cite{graham2019hover}. In contrast to the aforementioned bottom-up methods that achieve instance segmentation by extracting the pixel features and grouping them into nuclei instances, top-down methods~\cite{chen2020blendmask} directly predict the nuclei instances with a sliding window. In the field of nuclei analysis, several methods~\cite{zhou2019irnet,gong2021style} adopted the typical top-down instance segmentation method Mask-RCNN~\cite{he2017mask} that simultaneously predicts the nuclei instances and their corresponding masks. Such methods are especially suitable for overlapping nuclei. However, these methods are limited by the requirement of pixel-level annotations, which are time-consuming and expensive to obtain.

\subsection{Weakly-Supervised Image Segmentation}
\label{sec:related_work_weakly_supervised_segmentation}
Weakly-supervised semantic segmentation (WSSS) is a challenging task that aims to develop a model capable of predicting pixel-level labels using only weak supervision information to reduce manual efforts involved in training data labeling. Weak supervision information can be categorized into four categories, namely bounding box\cite{kulharia2020box2seg,pont2016multiscale,rother2012interactive}, scribble\cite{tang2018normalized}, point~\cite{qian2019weakly}, and image tags~\cite{ahn2018learning}, arranged from fine to coarse granularity. Typically, WSSS methods aim to incorporate domain knowledge into models to encompass the spatial and semantic information of the weak annotations. In particular, the auxiliary information can be categorized into priors and hints~\cite{zhang2021weakly,ouassit2022brief}, where priors refer to the prior knowledge of the data, independent of any specific sample, while hints refer to indirect information derived from the weak annotations. For instance, for priors, WSSS methods~\cite{kulharia2020box2seg} mainly employ the region proposals generated by MCG~\cite{pont2016multiscale} or GrabCut~\cite{rother2012interactive}, which are based on the concept that the foreground and background regions are separated by high-contrast boundaries. On the other hand, hints are used indirectly, as several methods~\cite{hsu2019weakly,wang2021bounding} have adopted multiple instance learning for WSSS with box annotation, where vertical and horizontal crossing lines are considered as the bag, and the pixels inside the line are treated as the instances.
However, in the task of nuclei segmentation in pathology images, the aforementioned methods are still limited by the high annotation cost. Compared to natural images, nuclei in pathology images are small and numerous, making manual annotation of nuclei instances in pathology images time-consuming and expensive. Additionally, it is also challenging to distinguish between irregular and touching nuclei instances. Therefore, obtaining all the nuclei instances in an image requires strong domain knowledge and manual efforts~\cite{qu2020weakly}. This paper aims to address this challenge through the most label-efficient WSSS method for nuclei segmentation (Fig.~\ref{fig:labels}), which only requires partial point annotations.

\subsection{Weakly-Supervised Nuclei Segmentation}
\label{sec:related_work_weakly_supervised_nuclei_segmentation}
Significant progress has been made recently in weakly-supervised nuclei segmentation in pathology images. 
To address class imbalance and overfitting problems, extra supervision information has been introduced, such as geometric diagrams and clustering labels~\cite{qu2019weakly,qu2020weakly,chamanzar2020weakly,yoo2019pseudoedgenet,tian2020weakly}. 
For example, Qu~\etal~\cite{qu2019weakly,qu2020weakly} proposed a weakly-supervised nuclei segmentation method that uses Voronoi diagrams~\cite{brown1979voronoi} and cluster labels~\cite{hartigan1979algorithm} to guide segmentation. 
It takes the Voronoi diagram as the prior information that the nuclei shapes are nearly convex and the point annotations are the centers of nuclei.
However, this assumption is not suitable for all nuclei instances, which may lead to inaccurate supervision information. Iterative refinement strategies~\cite{chamanzar2020weakly,tian2020weakly,qu2019weakly} have been proposed to improve models trained with coarse annotations (\ie, Voronoi and cluster) by replacing the Voronoi diagram and cluster labels with the prediction results.
Yoo~\etal~\cite{yoo2019pseudoedgenet} proposed a pseudo edge network that uses the pseudo edges generated by the Sobel filter to guide segmentation. 
To further exploit boundary information, Lin~\etal~\cite{lin2023nuclei} proposed a self-supervised method that makes the model implicitly aware of boundary information. 
However, these methods are limited by either inaccurate hand-crafted supervision or indirectly proxy tasks, resulting in sub-optimal segmentation performance. In this paper, we propose a novel boundary mining method that directly supervises the model to predict nuclei boundaries by learning pixel affinities between boundaries through a multiple instance learning objective.
 
\begin{figure*}
	\centering
	\includegraphics[width=\textwidth]{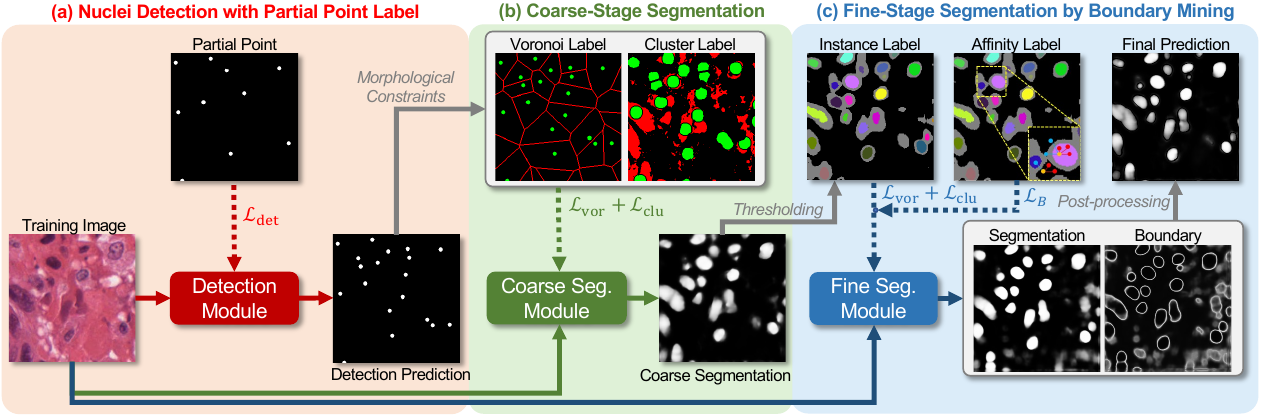}
	\caption{An overview of the proposed method. The nuclei detection module with partial point label is shown in (a). This is followed by the coarse-stage nuclei segmentation module in (b), which is supervised by Voronoi and cluster labels derived from the detection prediction in the previous step. Finally, the fine-stage segmentation module in (c) produces binary segmentation results that are supervised by the coarse segmentation results, while the boundary prediction is supervised by the affinity map in a multiple instance learning manner. The dashed lines denote the supervision information.}
	\label{fig:framework}
\end{figure*}

\section{METHODOLOGY}
\label{sec:method}
As illustrated in Fig.~\ref{fig:framework}, the proposed pipeline for nuclei instance segmentation mainly composes of three parts: 1) a nuclei detection module with partially labeled point annotation; 2) a nuclei segmentation module with the coarse labels derived from the point annotations; 3) a nuclei instance segmentation module that mines the boundary information from the former results. The details of each module are described as follows.

\subsection{Nuclei Detection with Partial Points}
\label{sec:det}
\subsubsection{Gaussian Masks for Point Annotations}
The first step uses the partially labeled point annotations to train a nuclei detection module. 
Typical methods for weakly-supervised instance segmentation of nuclei are based on the full annotations of nuclei center, which highly relies on the assumption that all the nuclei have been annotated and the point annotations are located at the center of the nuclei. 
However, these assumptions are not practical in real-world clinical settings. 
Therefore, this paper considers the partially labeled point annotations, which only require the pathologists to annotate part of the nuclei in each image, significantly reducing the annotation cost. 

To address the problem of class imbalance, we begin by transforming the point annotations into Gaussian heatmaps, which is a commonly used approach in keypoint detection tasks~\cite{pereira2022sleap}. 
The Gaussian heatmap is defined as:
\begin{equation}
\label{eq:heatmap}
\mathcal{H}_i  = 
\begin{cases}
\exp\left(-\frac{D_i^2}{2\sigma^2}\right)  & \text{if $D_i < r_1$,} \\
0  &  \text{if $r_1 < D_i < r_2$,} \\
-1 &  \text{otherwise,}	\end{cases}	
\end{equation}
where $\mathcal{H}_i$ is the $i$-th Gaussian heatmap, $\sigma$ is the standard deviation of the Gaussian kernel, $r_1$ and $r_2$ are pre-defined hyper-parameters indicating the radius of foreground and background, respectively.
For the pixel whose Euclidean distance is larger than $r_2$, we consider it as ignoring area, denoted as -1.
$D_i$ is the Euclidean distance between the $i$-th pixel and its closest point annotation, denoted as:
\begin{equation}
    D_i=\min\limits_{j\in N} (\|x_i-p_j\|^2),
\end{equation}
where $N$ is the total number of annotated points in the image.
The Gaussian heatmap is used to train the nuclei detection module, which serves as a pixel-wise regression task. We use the mean squared error as the loss function:
\begin{equation}
\label{eq:mse}
\mathcal{L_{\text{det}}} = \frac{1}{N}\sum_{\mathcal{H}_i\neq -1} (\mathcal{H}_i - \hat{\mathcal{H}}_i)^2,
\end{equation}
where $\hat{\mathcal{H}}_i$ is the predicted heatmap, $N$ is the total number of pixels that are not ignored in the image.
Considering the significant class imbalance problem between the annotated foreground point and the background area, we follow Qu~\etal~\cite{qu2020weakly} to set different weights for the foreground and background pixels, \ie, 1.0 and 0.1, respectively.

\subsubsection{Pseudo Label Selection with Curriculum Learning}
Drawing inspiration from curriculum learning~\cite{bengio2009curriculum}, we train the detection module in a single shot instead of iteratively training it with different confidence thresholds.
The general principle of curriculum learning is to train the model with easy samples first and then gradually increase the difficulty of the samples as the model improves. 
To determine the difficulty of each pixel, we make three key observations. 
First, we note that the shape and size of nuclei in histopathology images are relatively uniform compared to objects in natural images. 
Second, we find that pixels close to annotated points are easier to learn due to their more deterministic surroundings.
Lastly, we observe that pseudo foreground labels would tend to produce false positive areas larger than the normal nuclei. 
Based on these observations, we defined the training difficulty $\mathcal{TD}$ for each pixel:
\begin{equation}
    \label{eq:training_difficulty}
    \begin{split}
        \mathcal{TD}_i &= \mathcal{N}(\overline{D_i^k}) \times \mathcal{N}(A_i) \times \left( 1 - \mathcal{N}(S_i)\right), \\
    \end{split}
\end{equation}
where $\overline{D_i^k}$ is the mean distance to the $k$ nearest existing samples of the connected component of the $i$-th pixel, $A_i$ is the area of the connected component of the $i$-th pixel, and $S_i$ is the mean score predicted by the detection module of all pixels in the connected component of $i$-th pixel.
Specifically, we first binarize the probability map to generate  point label map with a threshold (set to 0.65), then compute the connected components for the binarized map.
$\mathcal{N}(\cdot)$ is the normalization function that maps the value to [0, 1].
We compute the training difficulty after every $t$ epochs ($t=30$ in our experiments) and then rank all samples that do not overlap existing labels from low to high according to their training difficulty.
The top samples will be added to the ground truth labels and supervise the subsequent training. To be specific, we control the number of added samples by:
\begin{equation}
    \label{eq:number_add}
    n_{\text{add}} =  n_{\text{det}} \times \exp(-\frac{n_{\text{gt}}}{n_{\text{det}}})
\end{equation}
where $n_{\text{det}}$ and $n_{\text{gt}}$ are the number of detected valid samples and the number of pseudo label samples.

\subsection{Coarse-Stage Nuclei Segmentation with Rough Annotations}
\label{sec:coarse_seg}
Once we have trained the nuclei detection module, we can generate predictions for nuclei detection across all entire images.
Note that even though the detection results may introduce some amount of noise labels for nuclei localization, they can still provide considerably more supervision for the segmentation task. 
Furthermore, deep neural networks are resilient to noise labels~\cite{zhang2022dtfd} to some extent.
Therefore, we can use the detection results as fully labeled point annotations to train the segmentation network.
Following typical weakly-supervised nuclei segmentation methods~\cite{qu2020weakly,qu2019weakly,tian2020weakly,lin2023nuclei,kostrykin2022superadditivity}, we transform point annotations into two coarse pixel-wise annotations, \ie, Voronoi and cluster annotations.

\subsubsection{Voronoi Annotations} 
A Voronoi diagram is a geometric construct that partitions a plane into distinct regions based on their distances to points in a specific subset of the plane.
Leveraging the observation that the nearly convex shapes of nuclei and the proximity of the point annotations to nuclei center, Voronoi diagram is a natural choice for generating coarse annotations from point annotations. 
Specifically, as illustrated in Fig.~\ref{fig:labels}(e), the image is partitioned into multiple cells using the Voronoi diagram, where each cell corresponds to a single-point annotation. 
The edges of the cells are the Voronoi boundaries, which are the perpendicular bisectors of the line segments connecting the center of the cell to the two nearest point annotations.
During the training process, the point annotations serve as the foreground labels, while the Voronoi boundaries serve as the background labels.

\subsubsection{Cluster Annotations}
To further propagate the Voronoi label to the entire image, we employ the $k$-means clustering algorithm~\cite{hartigan1979algorithm} to generate the cluster annotations.
Initially, we first generate a distance map from the point annotation, where the distance of each pixel to the nearest point annotation is computed.
Subsequently, combining the distance map and the original image, we use the $k$-means clustering algorithm to group the pixels into $k=3$ clusters, representing the foreground, background, and uncertain regions, as shown in Fig.~\ref{fig:labels}(f).
We use the cross-entropy loss function to train the coarse segmentation module with the Voronoi and cluster annotations as follows:
\begin{equation}
    \label{eq:loss_vor}
    \mathcal{L}_\text{vor} = \frac{1}{|\Omega_v|}\sum_{i\in\Omega_v} \left(y_i \log(\hat{y}_i) + (1-y_i) \log(1-\hat{y}_i)\right)
\end{equation}
\begin{equation}
    \label{eq:loss_clu}
    \mathcal{L}_\text{clu} = \frac{1}{|\Omega_c|}\sum_{i\in\Omega_c} \left(y_i \log(\hat{y}_i) + (1-y_i) \log(1-\hat{y}_i)\right)
\end{equation}
where $y_i$ and $\hat{y}_i$ are the ground truth and the prediction for the $i$-th pixel, and $\Omega_*$ (\ie, $\Omega_v$ and $\Omega_c$) is the set of pixels that are not ignored in the Voronoi or cluster label. 

\subsection{Fine-Stage Segmentation with Boundary Mining}
\label{sec:fine_seg}
Illustrated in Fig.~\ref{fig:labels}, the coarse pixel-wise annotations provide much more information compared with the original point annotations.
Nonetheless, the imprecise boundary details of the Voronoi and cluster labels engender a significant count of ambiguous pixels situated proximate to the nuclei instance borders, necessitating more informative guidance. 
To address this, we propose to utilize the pairwise pixel affinities to harness the boundary information derived from coarse annotations.

\subsubsection{Network Architecture}
\begin{figure}
	\centering
	\includegraphics[width=0.48\textwidth]{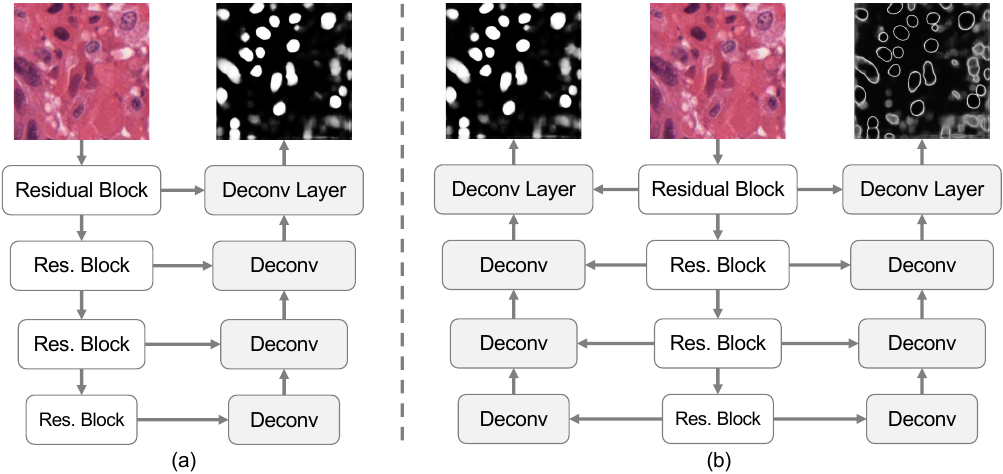}
	\caption{Architectures of (a) coarse-stage segmentation model; (b) fine-stage segmentation model.}
	\label{fig:finesegarch}
\end{figure}
As shown in Fig.~\ref{fig:finesegarch}(b), the fine-stage segmentation module is composed of two parts, \ie, the binary nuclei segmentation network and the boundary prediction network.
The two networks are both based on the U-Net Architecture~\cite{ronneberger2015u}, sharing the same backbone.
In this paper, we use the ResNet-34~\cite{he2016deep} as encoder, following the typical weakly-supervised nuclei segmentation methods~\cite{qu2019weakly,qu2020weakly,lin2023nuclei}.

\subsubsection{Learning Pixel Affinity from Coarse Annotations}
\label{sec:pixel_affinity}
\begin{figure}
	\centering
	\includegraphics[width=0.48\textwidth]{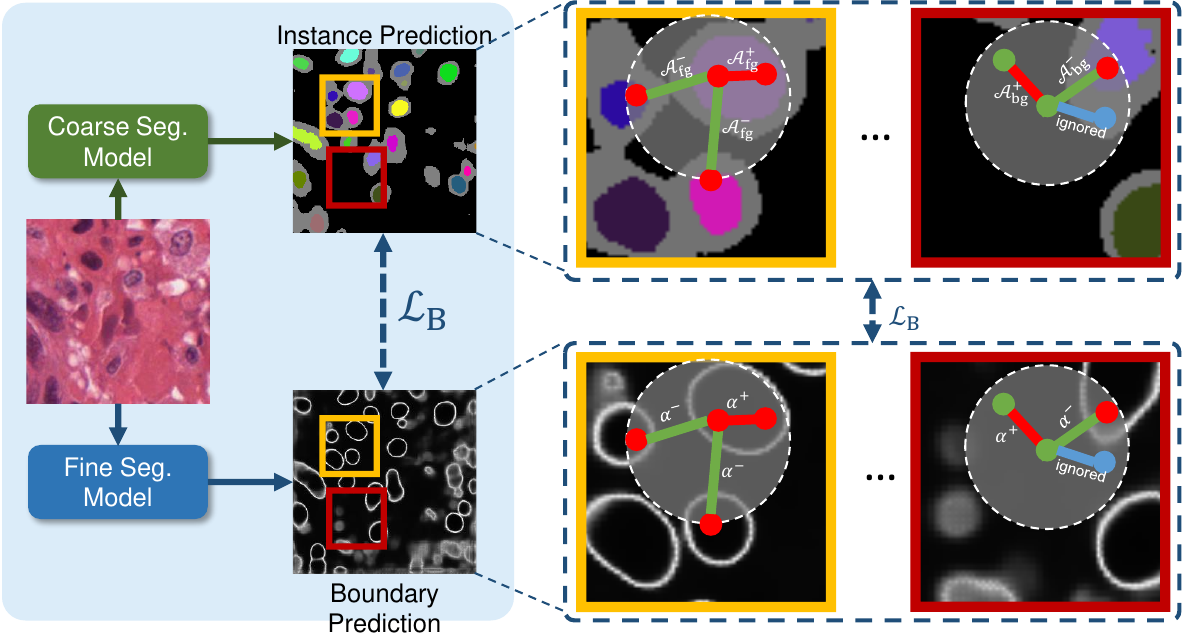}
	\caption{The boundary loss used in our method. The left panel shows how the nuclei instance prediction of the coarse model is used to supervise the boundary prediction of the fine-stage segmentation model. The right panel illustrates the calculation of the boundary loss using pairwise pixel affinity.}
	\label{fig:affinity}
\end{figure}

As shown in Fig.~\ref{fig:affinity}, the coarse annotations are used to generate the pixel affinity~\cite{ahn2019weakly,wang2020weakly}, which subsequently supervises the boundary prediction network.
Specifically, we begin by binarizing the coarse annotations using two distinct thresholds, namely $T_f$ and $T_b$. 
Pixels with predicted scores exceeding $T_f$ are considered foreground (colorful areas), while pixels with scores lower than $T_b$ are considered background (black areas). 
Pixels with predicted scores falling between these thresholds are considered uncertain (grey areas) and are ignored while labeling pixel affinity.
Note that the different colors of the foreground pixels indicate distinct nuclei, which not only provide the boundary information but also instance segmentation of the nuclei.
Then, we construct the pairwise pixel affinity map in accordance with the coarse annotations.
For each pixel situated within the confident areas (colorful and black), we sample its neighboring pixels, then determine the affinity of each pair of the pixel and its neighboring pixels within the circles of radius $\gamma$.
If the two pixels belong to the same instance, the affinity is set to 1, otherwise, the affinity is set to 0.
If any of the two pixels is uncertain, the affinity is set to -1, which will not be considered in the training process.
Formally, the affinity map is defined as follows:
\begin{equation}
    \label{eq:affinity}
    \mathcal{A}_{ij} = \left\{
    \begin{array}{ll}
    1, & \text{if } \mathbf{C}_{i} = \mathbf{C}_{j} \\
    0, & \text{if } \mathbf{C}_{i} \neq \mathbf{C}_{j} \\
    -1, & \text{otherwise},
    \end{array}
    \right.
\end{equation}
where $\mathbf{C}_{i}$ and $\mathbf{C}_{j}$ are the class of the $i$-th and $j$-th pixel, respectively.
Further, we divide the affinity set $\mathcal{A}$ of all pixel pairs into four subsets, \ie, $\mathcal{A}^{+}_{\mathbf{fg}}$, $\mathcal{A}^{-}_{\mathbf{fg}}$, $\mathcal{A}^{+}_{\mathbf{bg}}$, and $\mathcal{A}^{-}_{\mathbf{fb}}$, which denotes the affinity map of pixels in the same nucleus, pixels in different nuclei, pixels in the background, and the pairs where one of the pixels is in the foreground, and the other is in the background, respectively.

To supervise the boundary prediction network, we convert the boundary prediction into the affinity map.
Inspired by the multiple instance learning~\cite{zhang2022dtfd}, we construct the affinity map based on the premise that a pair of pixels with positive affinity indicates that the path connecting them will not intersect the boundary; otherwise, the path will cross the boundary.
Therefore, the affinity map can be derived from the boundary prediction map by the following equation:
\begin{equation}
    \label{eq:mil}
    a_{ij} = 1 - \max_{k\in \prod_{i,j}} \hat{y}_{k},
\end{equation}
where $y$ is the boundary prediction map, and $k\in \prod_{i,j}$ is the set of pixels on the path between the $i$-th and $j$-th pixel.
Based on the aforementioned assumption, if the path between the $i$-th and $j$-th pixel crosses the boundary, the max value of the boundary prediction map on the path $\prod_{i,j}$ should be close to 1, and the affinity of the pair of the $i$-th and $j$-th pixel will be close to 0.
We use the cross-entropy to train the boundary prediction network with the affinity map as follows:
\begin{equation}
    \label{eq:loss_b}
    \begin{aligned}
    \mathcal{L}_B = & -\sum_{(i,j)\in \mathcal{A}^{+}_{\mathbf{fg}}} \frac{\log(a_{ij})}{|\mathcal{A}^{+}_{\mathbf{fg}}|} - \sum_{(i,j)\in \mathcal{A}^{+}_{\mathbf{bg}}} \frac{\log(a_{ij})}{|\mathcal{A}^{+}_{\mathbf{bg}}|} \\
     &- \sum_{(i,j)\in \mathcal{A}^{-}_{\mathbf{fg}}} \frac{\log(1-a_{ij})}{|\mathcal{A}^{-}_{\mathbf{fg}}|}  - \sum_{(i,j)\in \mathcal{A}^{-}_{\mathbf{bg}}} \frac{\log(1-a_{ij})}{|\mathcal{A}^{-}_{\mathbf{bg}}|}
    \end{aligned}
\end{equation}
where the four terms in the equation are normalized by the number of the corresponding affinity set due to the class imbalance of the four subsets. 

The two branches of the fine-stage segmentation module are trained in an end-to-end manner, and the final loss function is defined as follows:
\begin{equation}
    \label{eq:loss_fine}
    \mathcal{L_{\text{fine}}} = \mathcal{L}_{\text{vor}} + \mathcal{L}_{\text{clu}} + \beta \mathcal{L}_B,
\end{equation}
where $\beta$ is the weight of the boundary loss, which will be discussed in Sec.~\ref{sec:abl}. We follow the previous work~\cite{qu2020weakly} that assign equal weight to the Voronoi loss and cluster loss.

\subsection{Inference}
Once the fine-stage segmentation module is trained, we can use a parameter-free post-processing approach to obtain the ultimate segmentation result. 
Specifically, we first perform boundary map subtraction from the segmentation map, followed by binarizing the resulting map using a constant threshold. 
We further apply morphological operations to fill small holes and remove small objects in the segmentation map. 
Then, the connected components are computed to obtain the final instance-level segmentation result. 
Finally, we empirically perform a dilation operation for each connected component using a disk structuring element of radius 1 pixel to obtain the definitive segmentation result.

\section{Experiments}
\label{sec:exp}

\subsection{Experiment Settings}
\subsubsection{Datasets}
We evaluate our method on three public datasets of H\&E stained histopathology images for nuclei segmentation, respectively, Multi-Organ Nuclei Segmentation (MoNuSeg)~\cite{monuseg_test}, Computational Precision Medicine (CPM)~\cite{cpm}, and Colon Nuclei Identification and Counting (CoNIC) Challenge dataset~\cite{graham2021conic}.
\begin{itemize}
\item \textbf{MoNuSeg} comprises 44 images of various patients diagnosed with tumors of diverse organs, captured at 40x magnification, with each of size $1000 \times 1000$ pixels. The dataset has 30/14 images for training/testing. We randomly split the training subset into a training set with 24 images and a validation set with 6 images.
\item \textbf{CPM} concists of 32 images of size 500 $\times$ 500 or $600 \times600 $ pixels at 40$\times$ magnification. For each set of 8 images of the same tumor type, we randomly designate one image for validation and two images for testing.
\item \textbf{CoNIC} consists of 4,981 image patches with a size of 256$\times$256. Following~\cite{graham2021conic,zhang2022deep,lin2023rethinking}, we randomly split all images into 7:1:2 for training, validation, and testing.
\end{itemize}
All the three datasets provide pixel-level nuclei segmentation annotations. Our point labels are randomly sampled from segmentation masks.
We keep the same training/validation/testing splits for comparison with other methods.

\begin{table*}[thbp]
    \centering
    \caption{Comparison (\%) with state-of-the-art methods on three datasets, where CS, FS, and ST denote coarse stage, fine stage, and self-training, respectively.}
    \label{tab:cmp_sota}
    \renewcommand\arraystretch{1.5}
    \setlength{\tabcolsep}{1pt}{
    \resizebox{\textwidth}{!}{
    \begin{tabular}{c|p{0.5cm}<{\centering}p{0.5cm}<{\centering}p{0.5cm}<{\centering}| c c c c c c c| c c c c c c c | c c c c c c c} 
    \multicolumn{4}{c}{} & \multicolumn{7}{c}{MoNuSeg~\cite{monuseg_test}} & \multicolumn{7}{c}{CPM~\cite{cpm}} & \multicolumn{7}{c}{CoNIC~\cite{graham2021conic}} \\
    \toprule
    \multicolumn{4}{c|}{Methods} & Acc.  & F1  & Dice  & AJI  & DQ  & SQ  & PQ  & Acc.  & F1  & Dice  & AJI  & DQ  & SQ  & PQ  & Acc.  & F1  & Dice  & AJI  & DQ  & SQ  & PQ  \\
    \midrule
    \multirow{7}{*}{\rotatebox{90}{SOTAs}} &\multicolumn{3}{l|}{Qu \etal\cite{qu2020weakly}} & \underline{91.52}	& 76.76	& 73.24	& 54.32 & 69.72	& 71.28	& 49.84 & 89.90 & 76.56 & 71.17 & 50.91 & 64.10	& 70.66	& 45.69 & 90.24 & \underline{69.20} & \underline{68.90} & \underline{49.73} & \textbf{67.87} & 68.06 & \underline{46.18} \\
    &\multicolumn{3}{l|}{Xie \etal\cite{xie2020instance}} & 91.19	& 77.56	& 72.51	& 51.69 & 68.82	& \underline{72.50}	& 49.97 & 89.96 & 75.87 & 70.28 & 49.40 & 59.49 & 70.87 & 42.68 & 87.33 & 67.5 & 65.98 & 43.04 & 62.42 & 67.15 & 44.08 \\
    &\multicolumn{3}{l|}{Cha. \etal\cite{chamanzar2020weakly}} & 91.04	& 74.18	& 71.70	& 53.69 & 69.40	& 69.84	& 48.75 & 88.57 & 70.69 & 66.44 & 45.92 & 57.06	& 67.80	& 39.20 & \underline{91.68} & 67.65 & 66.83 & 47.30 & 60.21 & \textbf{70.24} & 42.90 \\
    &\multicolumn{3}{l|}{Lee \etal \cite{lee2020scribble2label}} & 91.13	& 77.05	& 73.44	& 54.20 & 72.03	& 71.80	& 51.78 & 89.85 & 75.80 & 70.82 & 50.26 & 65.37	& 69.75	& 45.99 & 88.67 & 67.71 & 67.04 & 48.62 & 64.80 & 67.39 & 45.84 \\
    &\multicolumn{3}{l|}{Yoo \etal \cite{yoo2019pseudoedgenet}} & 86.51	& 72.11	& 56.68	& 29.03 & 43.97    & 72.21 & 31.86 & 90.50 & 79.81    & 72.46  & 49.42 & 61.04	& \underline{73.63}	& 45.10 & 88.21 & 50.78 & 44.98 & 25.76 & 11.40 & 56.54 & 6.74 \\
    &\multicolumn{3}{l|}{Tian \etal \cite{tian2020weakly}} & 88.01  & 71.47 & 63.96 & 40.51 & 50.67 & 67.19 & 34.12 & 87.87 & 71.74 & 64.39 & 42.11 & 42.86	& 63.03	& 27.06 & 84.00 & 49.49 & 33.18 & 13.25 & 6.72 & 59.24 & 4.00 \\
    &\multicolumn{3}{l|}{Lin \etal \cite{lin2023nuclei}} & 91.44    
    & \underline{77.64}
    & \underline{74.41}
    & 56.20
    & \underline{73.27}
    &{72.48}
    & \underline{53.19}
    &\underline{91.01}
    & \underline{79.97}
    &\underline{73.73}
    & 51.69 &
    \textbf{68.42}
    & 72.18
    &\underline{49.66} & \textbf{91.78} & 67.66 & 66.72 & 47.13 & 60.69 & \underline{69.35} & 42.59 \\
    
    \cmidrule{2-25}
    \multirow{7}{*}{\rotatebox{90}{Ours}} & CS & FS & ST & \\
    \cmidrule{2-25}
    
    & \checkmark & & & 90.95 & 73.96 & 72.89 & 55.23 &71.06 &72.09 &51.47 &
    89.79 & 75.51 & 70.62 & 50.95 & 64.65 & 70.26 & 46.01 & 87.97 & 66.58 & 65.60 & 46.88 & 61.02 & 67.51 & 43.38 \\
    
    & \checkmark & \checkmark & &
    \makecell[c]{91.18\\ \tiny$^{(\textcolor{red}{\uparrow0.23})}$} &
    \makecell[c]{75.59\\ \tiny$^{(\textcolor{red}{\uparrow1.63})}$} &
    \makecell[c]{73.85\\ \tiny$^{(\textcolor{red}{\uparrow0.96})}$} &
    \makecell[c]{\underline{57.40}\\ \tiny$^{(\textcolor{red}{\uparrow2.17})}$} & \makecell[c]{72.33\\ \tiny$^{(\textcolor{red}{\uparrow1.27})}$} &
    \makecell[c]{71.51\\ \tiny$^{(\textcolor{blue}{\downarrow 0.58})}$} &
    \makecell[c]{51.88\\ \tiny$^{(\textcolor{red}{\uparrow0.41})}$} &
    
    \makecell[c]{90.58 \\ \tiny$^{(\textcolor{red}{\uparrow 0.79})}$} &
    \makecell[c]{78.24\\ \tiny$^{(\textcolor{red}{\uparrow 2.73})}$} &
    \makecell[c]{73.33\\ \tiny$^{(\textcolor{red}{\uparrow 2.71})}$} &
    \makecell[c]{\underline{53.40}\\ \tiny$^{(\textcolor{red}{\uparrow 2.45})}$} &
    \makecell[c]{{64.10}\\ \tiny$^{(\textcolor{blue}{\downarrow 0.55})}$} &
    \makecell[c]{72.30\\ \tiny$^{(\textcolor{red}{\uparrow 2.04})}$} &
    \makecell[c]{46.61\\ \tiny$^{(\textcolor{red}{\uparrow 0.60})}$} &
    
    \makecell[c]{90.18\\ \tiny$^{(\textcolor{red}{\uparrow 2.21})}$} &
    \makecell[c]{68.16\\ \tiny$^{(\textcolor{red}{\uparrow 1.58})}$} &
    \makecell[c]{67.53\\ \tiny$^{(\textcolor{red}{\uparrow 1.93})}$} &
    \makecell[c]{50.32\\ \tiny$^{(\textcolor{red}{\uparrow 3.44})}$} &
    \makecell[c]{57.65\\ \tiny$^{(\textcolor{blue}{\downarrow 3.37})}$} &
    \makecell[c]{66.83\\ \tiny$^{(\textcolor{blue}{\downarrow 0.68})}$} &
    \makecell[c]{40.45\\ \tiny$^{(\textcolor{blue}{\downarrow 2.93})}$} \\
    
    & \checkmark & & \checkmark &
    \makecell[c]{90.91\\ \tiny$^{(\textcolor{blue}{\downarrow 0.04})}$} &
    \makecell[c]{72.50\\ \tiny$^{(\textcolor{blue}{\downarrow 1.46})}$} &
    \makecell[c]{71.78\\ \tiny$^{(\textcolor{blue}{\downarrow 1.11})}$} &
    \makecell[c]{54.39\\ \tiny$^{(\textcolor{blue}{\downarrow 0.84})}$} &
    \makecell[c]{69.82\\ \tiny$^{(\textcolor{blue}{\downarrow 1.24})}$} &
    \makecell[c]{70.51\\ \tiny$^{(\textcolor{blue}{\downarrow 1.58})}$} &
    \makecell[c]{49.50\\ \tiny$^{(\textcolor{blue}{\downarrow 1.97})}$} &
    
    \makecell[c]{89.60\\ \tiny$^{(\textcolor{blue}{\downarrow 0.19})}$} &
    \makecell[c]{74.75\\ \tiny$^{(\textcolor{blue}{\downarrow 0.76})}$} &
    \makecell[c]{70.38\\ \tiny$^{(\textcolor{blue}{\downarrow 0.24})}$} &
    \makecell[c]{50.88\\ \tiny$^{(\textcolor{blue}{\downarrow 0.07})}$} &
    \makecell[c]{58.87\\ \tiny$^{(\textcolor{blue}{\downarrow 5.78})}$} &
    \makecell[c]{69.41\\ \tiny$^{(\textcolor{blue}{\downarrow 0.85})}$} &
    \makecell[c]{41.21\\ \tiny$^{(\textcolor{blue}{\downarrow 4.80})}$} &
    
    \makecell[c]{88.68\\ \tiny$^{(\textcolor{red}{\uparrow 0.71})}$} &
    \makecell[c]{62.14\\ \tiny$^{(\textcolor{blue}{\downarrow 4.44})}$} &
    \makecell[c]{61.96\\ \tiny$^{(\textcolor{blue}{\downarrow 3.64})}$} &
    \makecell[c]{39.26\\ \tiny$^{(\textcolor{blue}{\downarrow 7.62})}$} &
    \makecell[c]{51.31\\ \tiny$^{(\textcolor{blue}{\downarrow 9.71})}$} &
    \makecell[c]{66.58\\ \tiny$^{(\textcolor{blue}{\downarrow 0.93})}$} &
    \makecell[c]{35.96\\ \tiny$^{(\textcolor{blue}{\downarrow 7.42})}$} \\
    
    & \checkmark & \checkmark & \checkmark &
    \makecell[c]{\textbf{91.81}\\ \tiny$^{(\textcolor{red}{\uparrow 0.86})}$} &
    \makecell[c]{\textbf{78.05}\\ \tiny$^{(\textcolor{red}{\uparrow 4.09})}$} &
    \makecell[c]{\textbf{76.73}\\ \tiny$^{(\textcolor{red}{\uparrow 3.84})}$} &
    \makecell[c]{\textbf{60.73}\\ \tiny$^{(\textcolor{red}{\uparrow 5.50})}$} &
    \makecell[c]{\textbf{75.59}\\ \tiny$^{(\textcolor{red}{\uparrow 4.53})}$} &
    \makecell[c]{\textbf{73.14}\\ \tiny$^{(\textcolor{red}{\uparrow 1.05})}$} &
    \makecell[c]{\textbf{55.43}\\ \tiny$^{(\textcolor{red}{\uparrow 3.96})}$} &
    
    \makecell[c]{\textbf{91.12}\\ \tiny$^{(\textcolor{red}{\uparrow 1.33})}$} &
    \makecell[c]{\textbf{79.99}\\ \tiny$^{(\textcolor{red}{\uparrow 4.48})}$} &
    \makecell[c]{\textbf{75.13}\\ \tiny$^{(\textcolor{red}{\uparrow 4.51})}$} &
    \makecell[c]{\textbf{54.54}\\ \tiny$^{(\textcolor{red}{\uparrow 3.59})}$} &
    \makecell[c]{\underline{66.96}\\ \tiny$^{(\textcolor{red}{\uparrow 2.31})}$} &
    \makecell[c]{\textbf{74.26}\\ \tiny$^{(\textcolor{red}{\uparrow 4.00})}$} &
    \makecell[c]{\textbf{49.91}\\ \tiny$^{(\textcolor{red}{\uparrow 3.90})}$} & 
    
    \makecell[c]{{91.14}\\ \tiny$^{(\textcolor{red}{\uparrow 3.17})}$} &
    \makecell[c]{\textbf{71.15}\\ \tiny$^{(\textcolor{red}{\uparrow 4.57})}$} &
    \makecell[c]{\textbf{70.33}\\ \tiny$^{(\textcolor{red}{\uparrow 4.73})}$} &
    \makecell[c]{\textbf{52.13}\\ \tiny$^{(\textcolor{red}{\uparrow 5.25})}$} &
    \makecell[c]{\underline{67.04}\\ \tiny$^{(\textcolor{red}{\uparrow 6.02})}$} &
    \makecell[c]{68.69\\ \tiny$^{(\textcolor{red}{\uparrow 1.18})}$} &
    \makecell[c]{\textbf{46.30}\\ \tiny$^{(\textcolor{red}{\uparrow 2.92})}$} \\ 	 	 	 	 	 
    \bottomrule
    \end{tabular}}}
    \end{table*}						
\begin{figure*}[htbp]
    \centering
    \vspace{-4mm}
    \includegraphics[width=\textwidth]{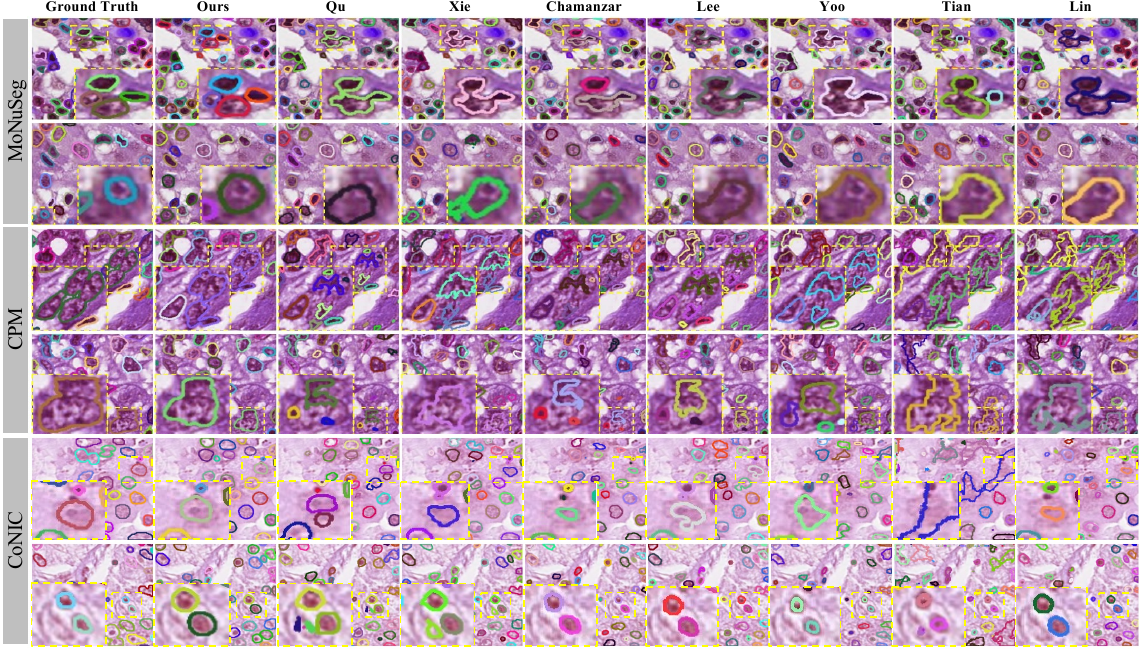}
    \caption{Visualization results on MoNuSeg (top two rows) and CPM (bottom two rows) datasets. Different colours of the nuclear boundaries denote separate instances. Regions of evident improvements are enlarged via yellow boxes to show better details.}
	\label{fig:visual}
    \vspace{-4mm}
\end{figure*}

\begin{figure*}[thbp]
	\centering
        \includegraphics[width=\textwidth]{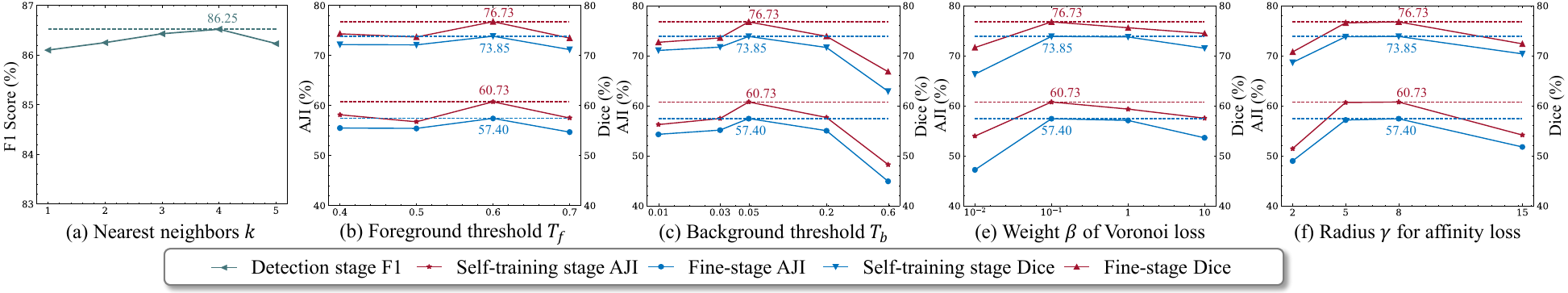}
	\caption{Ablation study on MoNuSeg dataset, where CS, FS, and ST denote coarse stage, fine stage, and self-training, respectively.}
	\label{fig:abl}
\end{figure*}

\subsubsection{Evaluation Metrics}
In the detection task, we adopt three evaluation metrics, \ie, precision, recall, and F1 score. 
The first two metrics determine the accuracy and sensitivity of the model, while F1 score reveals the overall detection quality.
As for the segmentation task, seven commonly accepted metrics are applied to evaluate our method, including accuracy, F1 score, object-level Dice coefficient (Dice)~\cite{dice}, aggregated Jaccard index (AJI)~\cite{monuseg_test}, detection quality (DQ), segmentation quality (SQ), and panoptic quality (PQ).
The first two metrics provide an assessment of the segmentation outcome at the pixel level, while the remaining five metrics offer a thorough analysis of the performance at the object level.

\subsubsection{Implementation Details}
We crop each training image into $250 \times 250$ patches with an overlap of 125 pixels. During training, each image is augmented by random cropping into $224 \times 224$ mini patches, random flipping, and random affine transformation. During testing, the images are split into $224 \times 224$ patches with an overlap of 80 pixels. In the detection stage, the network is trained with a fixed learning rate of $1\times10^{-4}$. In the segmentation stage, the networks are trained with an initial learning rate of $1\times10^{-3}$. Schedulers reduce the learning rate by a factor of 10 after every 20 epochs. We adopt Adam optimizer with a weight decay of $5\times10^{-4}$. The models with the lowest validation loss are selected for testing.
The proposed method is implemented using Pytorch and trained on an NVIDIA RTX 3090 GPU.
The inference time of our method is about 0.53 seconds per 1000$\times$1000 image.

\subsection{Results}
\subsubsection{Comparison with state-of-the-arts.}
In Table~\ref{tab:cmp_sota}, we compare the proposed BoNuS model to several state-of-the-arts (SOTAs) for nuclei segmentation with point annotation.
All the experiments are conducted with 100\% points in order to remove influence from the detection stage.
Our method outperforms other methods by a significant margin in terms of Dice, AJI and PQ, indicating its superior capability in identifying nuclei instances at the object level.
Notably, on the MoNuSeg dataset, our BoNuS model achieves 91.81\%, 78.05\%, 76.73\%, 60.73\%, 75.59\%, 73.14\%, and 55.43\%, in accuracy, F1 score, Dice, AJI, DQ, SQ, and PQ, respectively. 
On the CPM dataset, our BoNuS model achieves 91.12\%, 79.99\%, 75.13\%, 54.54\%, 66.96\%, 74.26\%, and 49.91\% in accuracy, F1 score, Dice, AJI, DQ, SQ, and PQ, respectively. 
On the CoNIC dataset, our BoNuS model achieves 91.14\%, 71.15\%, 70.33\%, 52.13\%, 67.04\%, 68.69\%, and 46.30\% in the above metrics.
Compared with other methods, our BoNuS model exhibits the best performance in most of metrics.
Especially in the AJI, our BoNuS model outperforms the second-best method by 4.53\%, 2.85\%, and 2.40\% on MoNuSeg, CPM, and CoNIC, respectively. 
The Visualization results are in Fig.~\ref{fig:visual}, which demonstrate the effectiveness of our BoNuS model in nuclei segmentation, especially at the object level.

\subsubsection{Component contributions.}
\begin{figure}[!t]
    \centering
    \includegraphics[width=0.48\textwidth]{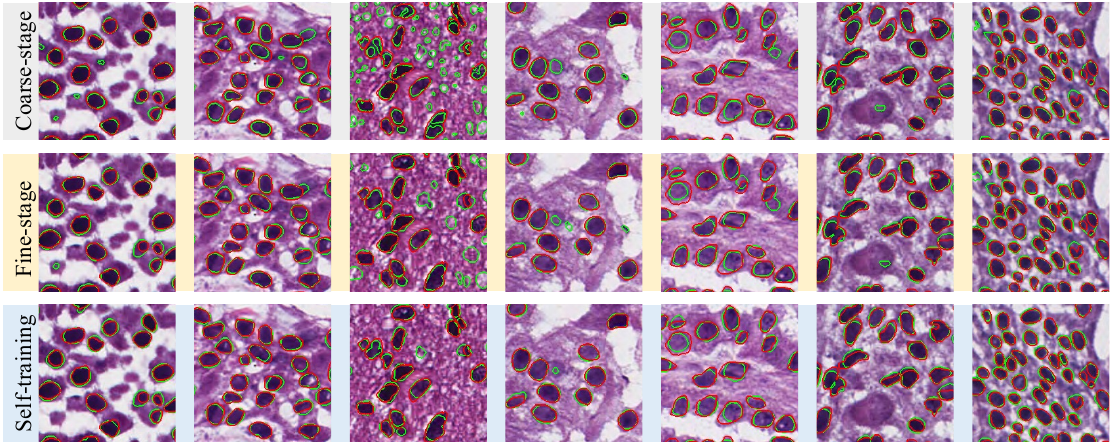}
    \caption{Visualization of the ablation study on MoNuSeg dataset. The red and green color indicates the ground-truth and prediction, respectively.}
	\label{fig:visual_abl}
\end{figure}

Table~\ref{tab:cmp_sota} presents a comparative analysis of the contributions made by individual components of the BoNuS model to the MoNuSeg and CPM datasets.
Our model comprises a coarse-stage segmentation module (CS) and a fine-stage segmentation module (FS), both of which are trained using distinct loss functions.
Specifically, CS is trained using Voronoi and cluster loss (refer to Sec.~\ref{sec:coarse_seg}), while FS is trained using interior and boundary loss, which are generated by pseudo labels created by CS (refer to Sec.~\ref{sec:fine_seg}).
Self-training (ST) is performed by retraining the segmentation network in a fully supervised learning manner with pseudo labels generated by the former weakly supervised segmentation network, which is a simple yet effective procedure in WSSS~\cite{zhang2020survey}.
On the MoNuseg dataset, CS achieves 90.95\%, 73.96\%, 72.89\%, 55.23\%, 71.06\%, 72.09\%, and 51.47\% in accuracy, F1 score, Dice, AJI, DQ, SQ, and PQ, respectively.
With FS, the corresponding absolute performance improvements are 0.23\%, 4.09\%, 1.96\%, 2.17\%, 1.27\%, -0.58\%, and 0.41\%, respectively.
Interestingly, ST has the opposite effect on CS and FS.
Specifically, for CS, ST gets slight performance drop by -0.04\%, -1.46\%, -1.11\%, -0.84\%, -1.24\%, -1.58\%, and -1.97\%, in accuracy, F1 score, Dice, AJI, DQ, SQ, and PQ, respectively.
Contrastly, for FS, ST achieves significant performance improvements by 0.86\%, 4.09\%, 3.84\%, and 5.50\%, 4.53\%, 1.05\%, and 3.96\% in accuracy, F1 score, Dice, AJI, DQ, SQ, and PQ, respectively.
This is because the pseudo labels generated by CS lack accuracy, particularly for boundary regions, leading to the self-deception problem, \ie, the false positive predictions are amplified by ST.
The qualitative results are shown in Fig.~\ref{fig:visual_abl}, which demonstrate the effectiveness of each component in our BoNuS model. The similar trend can be found on the CPM and CoNIC datasets.

\subsection{Ablation Study}
\label{sec:abl}
During the nuclei detection phase, the number of neighbors $k$ in Eq.~(\ref{eq:training_difficulty}) is a critical parameter.
To study the effect of $k$, we vary it from 1 to 10.
We report the F1 score of the nuclei detection module on the MoNuSeg dataset in Fig.~\ref{fig:abl}(a).
The detection results indicate that the detection performance reaches its peak when $k$ is 4.
In the pixel-affinity map (Sec.~\ref{sec:pixel_affinity}), the thresholds $T_f$ and $T_b$ balance the ratio of foreground, background, and ignored area. 
Higher $T_f$ and lower $T_b$ would produce less while preciser foreground and background area, respectively, yielding more ignored area.
By reporting the AJI and Dice in Fig.~\ref{fig:abl}(b) and Fig.~\ref{fig:abl}(c), we show that the optimal performance is achieved when $T_f$ and $T_b$ are 0.6 and 0.05, respectively, on the MoNuSeg dataset.
The hybrid loss consists of a critical weighting factor $\beta$, which determines the balance between the interior segmentation loss (\ie, Voronoi loss and cluster loss) and the boundary loss.
We assign the equal weight to the Voronoi loss and cluster loss following~\cite{qu2020weakly} and report the fine-stage segmentation results in AJI versus different $\beta$ on the MoNuSeg dataset in Fig.~\ref{fig:abl}(d).
The results show that the segmentation models achieve the best performance when $\beta$ is 0.1.
In the boundary loss, the radius $\gamma$ that limits the maximum distance of the pairwise pixel is a critical parameter, which is used to determine the boundary region.
To investigate the impact of $\gamma$, we set it to 2, 5, 8, 15, respectively, and report the AJI results of the fine-stage segmentation module on the MoNuSeg dataset in Fig.~\ref{fig:abl}(f).
Obviously, the segmentation results achieve the best performance when $\gamma$ is 8.
We argue that these parameters could be different for different datasets, which is related to the size and density of the objects.
Notably, our method demonstrates robustness under different settings, as it shows less sensitivity to variations in parameters.
This highlights the versatility and effectiveness of our approach for a wide range of datasets. 

\begin{figure}[!t]
    \centering
    \includegraphics[width=0.48\textwidth]{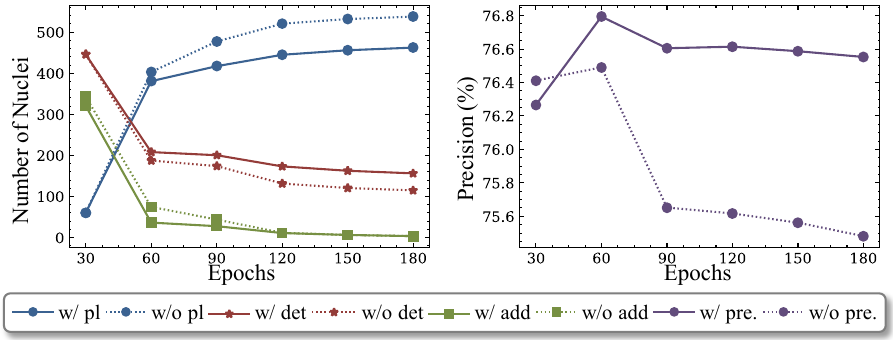}
    \caption{The number of nuclei used for training w/ and w/o curriculum learning. With curriculum learning, the pseudo labels generated are more precise and facilitate easier learning.}
	\label{fig:det}
\end{figure}

In Fig.~\ref{fig:det}, we perform a qualitative analysis of the pseudo labels generated by the nuclei detection module. We can observe that with the curriculum learning, the updated pseudo labels would be more precious, facilitating the learning process.
\begin{table}[!t]
    \centering
    \caption{Detection Results On MoNuSeg Dataset}
    \renewcommand\arraystretch{1.2}
    \begin{tabular}{c|ccc|ccc}
    \multicolumn{1}{c}{}  & \multicolumn{3}{c}{Ours} & \multicolumn{3}{c}{Qu~\etal~\cite{qu2020weakly}} \\
    \toprule
    Ratio & Precision & Recall & F1 & Precision & Recall & F1 \\
    \midrule
    0.05 & 85.35 & 85.37 & 85.36 & 85.12 & 83.44 & 83.76 \\
    
    0.10 & 87.26 & 86.40 & 86.82 & 85.28 & 86.61 & 85.94 \\

    0.25 & 87.78 & 85.13 & 86.43 & 89.99 & 81.98 & 85.79 \\

    0.50 & 87.40 & 85.86 & 86.62 & 83.72 & 87.61 & 85.62 \\

    0.75 & 86.61 & 85.64 & 86.12 & 82.46 & 82.25 & 82.35 \\
    
    Full & 87.77 & 87.22 & 87.49 & 82.84 & 85.55 & 84.17 \\
    \bottomrule
    \end{tabular}
    \label{tab:detection}
\end{table}

\subsection{Performance Versus Different Annotation Costs}
In this section, we study the performance of our BoNuS model with different annotation costs.
We report the results of detection and segmentation on the MoNuSeg dataset.
\subsubsection{Detection with Partial Points}
In Table~\ref{tab:detection}, we report the detection results with different annotation costs.
Our detection model achieves 87.49\% F1 score with 100\% point annotation. Remarkably, even with only 10\% point annotation, our detection model records 86.82\% F1 score, with a trivial 0.67\% performance drop, demonstrating the robustness of our detection model. Our proposed curriculum learning strategy consistently outperforms the baseline model across different ratios of point annotation, especially when the ratio is low.

\subsubsection{Segmentation with Partial Points}
\begin{figure}[htbp]
    \centering
    \includegraphics[width=0.48\textwidth]{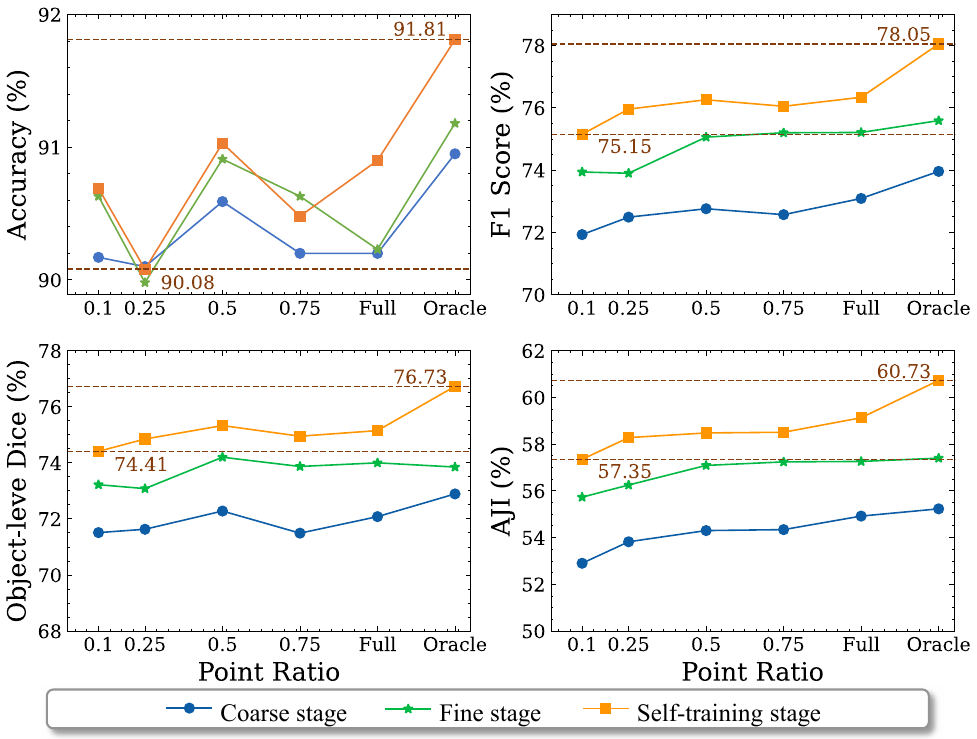}
	\caption{Model performance using different ratios of point annotation. \textit{Oracle} denotes that using the full point annotations, \textit{Full} means the using the detection results of model trained with full point annotations.}
	\label{fig:ratio}
\end{figure}
In Fig.~\ref{fig:ratio}, we report the segmentation results with different annotation costs.
We evaluate three components of our BoNuS model, \ie, coarse-stage segmentation module (CS), fine-stage segmentation module (FS), and self-training module (ST).
The results show that all three components achieve robust performance with different ratios of point annotation.
Especially for FS $+$ ST, with only 10\% point annotations, the segmentation results achieve 90.63\%, 73.94\%, 73.22\%, and 55.73\% in accuracy, F1 score, Dice, and AJI, respectively, obtaining only 0.27\%, 2.50\%, 1.67\%, and 3.40\% performance drop compared with the 100\% point annotation case.
Moreover, with only 10\% point annotations, the proposed BoNuS method outperforms the baseline model (\ie, CS $+$ ST) even when annotated with 100\% point, which emphasizes the effectiveness of our FS module.
This finding highlights the effectiveness and robustness of our proposed self-training strategy.
However, it should be noted that the performance of our BoNuS model is still limited by the detection performance, which is also the bottleneck of the whole pipeline.
Specifically, with 100\% point annotations, compared with the segmentation results that use the ground truth point annotation (\ie, \textit{Oracle}), the segmentation results that use the detection results (\ie, \textit{Full}) obtain 0.91\%, 1.71\%, 1.58\%, and 1.60\% performance drop in accuracy, F1 score, Dice, and AJI, respectively.

 
\section{Conclusion}
\label{sec:con}
In this paper, we investigate a fundamental yet pivotal problem in pathology image analysis, nuclei segmentation.
We consider the most challenging scenario where only partial point annotation is available, and propose a novel boundary mining framework, BoNuS, to learn the boundary information from the point annotation.
We first propose a novel nuclei detection method to generate the pseudo point labels for all the nuclei in the image.
Our nuclei detection method integrates the domain-specific prior knowledge of nuclei into the model via curriculum learning, significantly improving the detection performance, especially with limited point annotation.
Then, we propose a novel boundary mining loss to facilitate explicit learning of boundary information from the point annotation.
To address the issue of missing boundary supervision in the training process, we introduce a novel boundary mining loss that uses the pairwise pixel affinities to guide the model to learn the boundary information in a multiple instance learning formulation.
We conduct extensive evaluations on two public datasets, namely the MoNuSeg and CPM datasets, and demonstrate the effectiveness of our proposed approach. 
\bibliographystyle{IEEEtran}
\bibliography{ref}
\end{document}